\documentclass[acmsmall, manuscript]{acmart}
\pdfoutput=1

\usepackage{graphicx}
\usepackage{subfigure}

\usepackage{amssymb} 
\usepackage{amsthm}
\usepackage{algorithm}
\usepackage{multirow}
\usepackage{bm}
\usepackage{color}
\usepackage{listings}
\usepackage{algpseudocode} 

\usepackage{color}

\newcommand{\tabincell}[2]{\begin{tabular}{@{}#1@{}}#2\end{tabular}}

\AtBeginDocument{%
\providecommand\BibTeX{{%
		\normalfont B\kern-0.5em{\scshape i\kern-0.25em b}\kern-0.8em\TeX}}}

\setcopyright{acmcopyright}
\copyrightyear{2018}
\acmYear{2018}
\acmDOI{10.1145/1122445.1122456}

\acmJournal{JACM}
\acmVolume{37}
\acmNumber{4}
\acmArticle{111}
\acmMonth{8}

\begin{document}


\title{Exploiting Heterogeneous Graph Neural Networks with Latent Worker/Task Correlation Information for Label Aggregation in Crowdsourcing}


\author{Hanlu Wu}
\affiliation{%
	\institution{Zhejiang University}
	\country{China}}
\email{wuhanlu@zju.edu.cn}

\author{Tengfei Ma}
\affiliation{%
	\institution{IBM T. J. Watson Research Center}
	\country{USA}}
\email{tengfei.ma1@ibm.com}

\author{Lingfei Wu}
\affiliation{%
	\institution{JD.COM Silicon Valley Research Center}
	\country{USA}}
\email{lwu@email.wm.edu}

\author{Fangli Xu}
\affiliation{%
  \institution{Squirrel AI Learning}
  \country{USA}
}
\email{lili@yixue.us}

\author{Shouling Ji}
\authornote{Shouling Ji is the corresponding author.}
\affiliation{%
	\institution{Zhejiang University}
	\country{China}}
\email{sji@zju.edu.com}


\begin{abstract}
	Crowdsourcing has attracted much attention for its convenience to collect labels from non-expert workers instead of experts. However, due to the high level of noise from the non-experts, a label aggregation model that infers the true label from noisy crowdsourced labels is required. In this paper, we propose a novel framework based on graph neural networks for aggregating crowd labels. We construct a heterogeneous graph between workers and tasks and derive a new graph neural network to learn the representations of nodes and the true labels. Besides, we exploit the unknown latent interaction between the same type of nodes (workers or tasks) by adding a homogeneous attention layer in the graph neural networks. Experimental results on 13 real-world datasets show superior performance over state-of-the-art models.
\end{abstract}



\thanks{This work was partly supported by NSFC under No.61772466, U1936215, and U1836202, the National Key Research and Development Program of China under No.2020YFB2103802, 2018YFB0804102, and 2020AAA0140004, the Zhejiang Provincial Natural Science Foundation for Distinguished Young Scholars under No.LR19F020003, and the Fundamental Research Funds for the Central Universities (Zhejiang University NGICS Platform).}

\maketitle

\section{Introduction}

Recent years have witnessed the successful applications of machine learning in many fields, such as computer vision and natural language processing. Most of the machine learning tasks require large amounts of labeled data, however, obtaining labeled data from experts is quite expensive and time-consuming. Therefore, crowdsourcing has flourished as one of the most important tools for data annotation and labeling. With online platforms such as Amazon Mechanical Turk (AMT) \footnote{Amazon Mechanical Turk (AMT) can be found in www.mturk.com} and CrowdFlower~\footnote{CrowdFlower can be found in www.crowdflower.com}, one can get efficient and inexpensive access to crowdsourced resources.

Crowdsourcing systems generally break down a time-consuming task into more manageable microtasks, which can then be accomplished by distributed workforce independently. For example, to obtain enough labels for training a classifier, one can break down the labeling task into microtasks by assigning non-overlapping items to different workers for annotation. However, this way of task assignment suffers from the unprofessional nature of the workers, which leads to highly noisy data. A common practice is to increase the overlap of assignments between workers, i.e., collecting multiple labels from different workers for each single item. Then the ground-truth label can be induced from the noisy crowdsourced labels. By aggregating the wisdom of crowds, one can reduce the error rates and thereby improve the quality of the labeled data \cite{surowiecki2005wisdom}.

An intuitive strategy for crowdsourcing label aggregation is majority voting \cite{nitzan1982optimal}. However, this simple strategy is deficient for it ignores some important factors, such as worker ability. In a crowdsourcing system, workers usually show different expertise or reliability within a certain task, and a worker may be acquainted with some sort of tasks but fail when facing some others. What's worse is that a malicious worker may even submit wrong answers intentionally. As a consequence, majority voting, which equally treats each worker, can hardly make a reliable enough inference.

If the exact ability of each worker is known, the ground-truth label can be better inferred by weighted majority voting~\cite{littlestone1989weighted}. Based on this assumption, \citeauthor{tao2018domain} managed to learn the weight of each worker for weighted majority voting~\cite{tao2018domain}. However, we believe that the abilities of workers cannot be simply quantified by a single value. A worker may have a relatively strong ability in labeling one type of items correctly, but not good at another type of items (e.g. some workers are familiar with road signs and thus more professional in labeling related items,  but they know little about animals and are easy to make mistakes in distinguishing animals.).  Findings from existing work on crowdsourcing illustrate that it is crucial to model multi-dimensional latent features of workers~\cite{dawid1979maximum,whitehill2009whose,welinder2010multidimensional,tao2018domain}, which indicates different aspects of the workers. Meanwhile, the latent features of items (in the following, an item to be labeled is also called as a \textit{task}) also count for a lot. The difficulty of a task impacts the average rating as well as each worker's ability. 
Various probabilistic models have been proposed under the assumption that worker abilities and task difficulties are both key factors for inferring true labels~\cite{whitehill2009whose,yin2017aggregating} and obtained performance superior to majority voting. However, most of them require a delicate design for a sophisticated generative process and complex inference algorithms, and they are difficult to be generalized to large-scale datasets. Besides, there are also some deep learning models that jointly learn a classifier together with the label aggregation model~\cite{7405343,rodrigues2018deep,cao2018max,chu2020learning}. However, they usually require given features for each task, and different feature extraction strategies or different model structures of the classifier are needed according to the labeling tasks. Hence, the necessity of input task features will reduce the applicability of the model to some extent. In contrast, in this paper we are assumed to know only the task assignments of workers and their labels, and the workers and tasks are simply identified by ID numbers.

In order to model the relationship between workers and tasks, we propose to apply heterogeneous graph neural networks to crowdsourced label aggregation. To construct the graph, we model workers and tasks as two different types of nodes. If a worker and a task is connected by an edge, it indicates that the task was labeled by the worker. The main idea of graph neural networks is to iteratively aggregate information from their local neighborhoods, thus the graph neural network can naturally model the mutual interaction between tasks and workers and learn a good representation for them. We then infer the true label of a task from its representation. In this way, the crowdsourcing label inference problem is turned into a node classification problem in graph neural networks. 

Despite the representation power of graph neural networks, in our constructed graphs they can only utilize the assignment relationship between workers and tasks, while ignoring the workers' or tasks' latent relationship. Workers' correlation has been identified as another important factor for increasing the truth label inference in crowdsourcing~\cite{li2019exploiting}. Motivated by this observation, we further take into account the latent worker correlation, as well as task correlation, in our model and develop a new heterogeneous graph neural network based framework for crowdsourcing. In addition to the message passing between worker nodes and task nodes, we build an extra layer to implicitly propagate information among the same type of nodes, which has never been explored by previous heterogeneous graph neural networks to the best of our knowledge. 

Our contributions are summarized as follows:
\begin{itemize}
	
	\item We provide a new perspective for crowd label aggregation in the context of graph representation learning. To the best of our knowledge, it is the first model utilizing graph neural network to solve the crowdsourcing problem.
	
	\item Different from existing heterogeneous graph neural networks and most crowd label aggregation methods, our model learns a latent interaction among the same type of nodes to implicitly integrate the worker correlation and task correlation.
	
	\item We experiment on 13 real-world crowdsourcing datasets and demonstrate advantageous performance over state-of-the-art models. We also conduct ablation studies to explain the effectiveness of different components.
\end{itemize}

\section{Related Work}
\subsection{Crowdsourcing}

The increasing popularity of crowdsourcing as a labeling tool has led to a lot of attention to solve the issues of noisy crowdsourced labels. The early work of label aggregation can be traced back to ~\cite{dawid1979maximum}, which firstly proposed an Expectation-Maximization(EM)-based model to estimate the error rate of patients' answers to clinical problems. This model can be naturally transferred to the label aggregation problem. It utilizes workers' latent aspects by using a confusion matrix indicating the probability of a worker to choose each label for a task given the true label of it. 

Many follow-up studies can be viewed as extensions of the Dawid \& Skene model \cite{whitehill2009whose,liu2012variational, zhou2012learning, venanzi2014community, tian2015max, Khetan2016Achieving, yin2017aggregating}. Some work introduced task heterogeneity. In~\cite{zhou2012learning}, the authors incorporated both abilities and difficulties for workers and tasks respectively and inferred the truth using a min-max entropy principle. \citeauthor{venanzi2014community} modeled workers in community clusters to make workers share similar confusion matrices within the community~\cite{venanzi2014community}.~\citeauthor{Khetan2016Achieving} also introduced task difficulty into the Dawid \& Skene model and designed an adaptive task assignment scheme to provide more budget for tasks with more difficulty~\cite{Khetan2016Achieving}. The GLAD model (Generative model of Labels, Abilities, and Difficulties) considered both the abilities of workers and the difficulties of tasks and can simultaneously infer true labels as well as worker ability and task difficulty~\cite{whitehill2009whose}. LAA (Label-Aware Autoencoders) trains a classifier and a reconstructor, and the truth is inferred by the classifier as latent features~\cite{yin2017aggregating}. They also provided two extended models in their paper by considering object ambiguity (LAA-O) or latent aspects (LAA-L).
From the above-mentioned work, we can safely draw a conclusion that it's necessary to model the heterogeneity of both workers and tasks. Table~\ref{tab:method_comparison} compares a few methods in task modeling, worker modeling and correlation modeling (part of this table is quoted from \cite{zheng2017truth}).
Different from previous methods, EBCC (enhanced Bayesian classifier combination) additionally captures worker-worker correlations by dividing each true class into several subtypes and modeling the correlations between workers in the subtype level. Their approach infers true labels using a mean-field variational approach~\cite{li2019exploiting}. Inspired by this work, our model also incorporates inner-worker correlation. However, we also model the inner-task correlation in addition.

Other methods have been explored to select workers who can produce high-quality labels.
Based on the assumptions that some workers may assign labels casually (these workers are called \textit{spammer}), \citeauthor{eliminating} defined a spammer score to rank the workers and proposed an empirical Bayesian algorithm to iteratively eliminate the workers with high spammer score and estimate the ground-truth labels based only on those with low spammer score \cite{eliminating}.  
\citeauthor{ipeirotis2010quality} tried to evaluate the score of workers before task assignment and only assign tasks to workers with higher scores \cite{ipeirotis2010quality}. 
CrowdDQS dynamically issues golden standard questions and estimate the accuracies of workers in real-time, then it can select workers with higher accuracies for task assignment \cite{khan2017crowddqs}. \citeauthor{tu2020attention} suggest that the attention of workers changes over time, thus the accuracy of workers can not be kept constant, therefore, they proposed a probabilistic model that takes into account workers' attention \cite{tu2020attention}. Compared to these models, this paper focuses on a different scenario and our assumption is that the ability of a worker is diverse but constant (i.e. a worker will always give the same label to the same task).

\begin{table}[h]
	\centering
	\caption{Comparisons of Existing Methods. "$\times$" indicates the model does not consider this aspect.}
	\label{tab:method_comparison}
	\scalebox{1}{
		\begin{tabular}{|l|c|c|c|c|c|} 
			\hline
			\textbf{Method} & \textbf{Task} & \textbf{Worker} &\textbf{Worker-Worker Corr} &\textbf{Task-Task Corr} &\textbf{Worker-Task Corr}\\
			\cline{1-6}
			MV &$\times$ &$\times$ &$\times$ &$\times$ &$\times$ \\ \cline{1-6}
			D\&S~\cite{dawid1979maximum} &$\times$ &\checkmark  &$\times$ &$\times$ &$\times$ \\ \cline{1-6}
			ZC~\cite{demartini2012zencrowd} &$\times$ & \checkmark &$\times$ &$\times$ &$\times$\\\cline{1-6}
			Minimax~\cite{zhou2012learning} &$\times$ & \checkmark &$\times$ &$\times$ &\checkmark\\\cline{1-6}
			GLAD~\cite{whitehill2009whose} &\checkmark & \checkmark &$\times$ &$\times$ &$\times$ \\\cline{1-6} 
			BCC~\cite{kim2012bayesian} &$\times$ &\checkmark &$\times$ &$\times$ &$\times$ \\\cline{1-6} 
			LFC~\cite{raykar2010learning} &$\times$ &\checkmark &$\times$ &$\times$ &$\times$\\\cline{1-6}
			iBCC-MF~\cite{li2019exploiting} &$\times$ &\checkmark &$\times$ &$\times$ &$\times$\\\cline{1-6}
			EBCC~\cite{li2019exploiting} &$\times$ &\checkmark  &\checkmark &$\times$ &$\times$\\\cline{1-6}
			LAA~\cite{yin2017aggregating} &\checkmark &\checkmark &$\times$ &$\times$ &$\times$ \\\cline{1-6}
			CATD~\cite{li2014confidence} &$\times$ &\checkmark &$\times$ &$\times$ &$\times$\\\cline{1-6}
			PM~\cite{aydin2014crowdsourcing,li2014resolving} &$\times$ &\checkmark  &$\times$ &$\times$ &$\times$\\\cline{1-6}
			The proposed  &\checkmark  &\checkmark &\checkmark &\checkmark &\checkmark \\
			\hline
		\end{tabular}
	}
\end{table}

\subsection{Graph Neural Networks and General Frameworks}

A graph is a structured data consisting of nodes and edges connecting them. Data in many application scenarios has a natural graph structure, such as social networks, molecular structures, etc. In these scenarios, traditional deep learning methods are difficult to apply to the graph data. Therefore, in recent years, there is increasing interest in extending deep learning algorithms to the field of graphs as Graph Neural Networks (GNNs)~\cite{scarselli2008graph,kipf2016semi,hamilton2017inductive,chen2020iterative}. GNNs are capable of dealing with non-Euclidean structured data such as protein interaction networks~\cite{zitnik2018modeling}, citation networks~\cite{kipf2016semi}, traffic networks~\cite{lv2020temporal}, social networks, knowledge graphs~\cite{Hamaguchi2017Knowledge,gao2020rdf},  device-sharing network~\cite{liangstole,Liu2018GeniePath}, and text graph in natural language processing~\cite{chen2019reinforcement} etc. 

Some of these scenarios have various types of entities and relations (i.e. nodes and edges in the graph), hence called heterogeneous graphs. 
Several heterogeneous graph neural networks have been proposed and applied to various domains recently~\cite{chen2020toward,zhang2018deep,chen2018heterogeneous,wang2019heterogeneous}. 
To illustrate some,~\citeauthor{zitnik2018modeling} developed a heterogeneous graph neural network for drug side effect detection~\cite{zitnik2018modeling};~\citeauthor{fan2019graph} used heterogeneous graph neural networks for product recommendation~\cite{fan2019graph};~\citeauthor{wang2019heterogeneous} proposed a heterogeneous graph neural network with hierarchical attention mechanism that aggregates information from meta-path based neighbors \cite{wang2019heterogeneous}.
To the best of our knowledge, our work is the first trial to combine graph neural networks with the label aggregation problem in crowdsourcing. Moreover, different from previous heterogeneous graph neural networks, our work is the first one modeling the \textbf{implicit correlation} among the same type of nodes in a heterogeneous graph.

Some studies on general frameworks for graph neural networks have also emerged \cite{zhou2018graph,gilmer2017neural,wang2018non,battaglia2018relational}. \citeauthor{gilmer2017neural} proposed message passing neural network (MPNN) which unified various graph neural network approaches \cite{gilmer2017neural}. MPNN abstracts these graph neural networks into two phases, message passing phase and readout phase. The message passing phase aggregates information from the neighborhood based on a message function and an update function, and the readout phase is to obtain a representation of the whole graph based on the hidden states of each node. Our model is designed under MPNN framework. \citeauthor{wang2018non} proposed non-local neural network (NLNN) to capture the non-local dependencies of nodes \cite{wang2018non}. \citeauthor{battaglia2018relational} unified most of the graph neural networks including MPNN and NLNN by a graph networks (GN) framework \cite{battaglia2018relational}.

\begin{table}
	\centering
	\caption{Notation and Explanation}
	\label{notation}
	\begin{tabular}{|c|l|}
		\hline
		Notation & Definitions and Description\\
		\hline
		$\bm{u_i}$   					& worker node $i$ \\
		\hline
		$\bm{v_j}$   					& task node $j$ \\
		\hline
		$n$                            & number of workers \\
		\hline
		$m$                            & number of tasks \\
		\hline
		$g_j$   					&the label of task $j$
		inferred using majority voting\\
		\hline
		$l_{ij}$ 					    & crowd label given to task $j$ by worker $i$\\
		\hline
		$e_{ij}$ 						& \tabincell{l}{a one-hot vector indicating the crowd label given\\ to task $j$ by worker $i$}\\
		\hline
		$\mathcal{N}(i)$               & neighborhood of node $i$\\
		\hline
		$\mathcal{N}(\bm{u}_i)$  		& the set of tasks labeled by worker $\bm{u}_i$\\
		\hline
		$\mathcal{C}(\bm{v}_j)$ 		& the set of workers assigning labels to task $\bm{v}_j$ \\
		\hline
		$\bm{h}_i^{t} $                & hidden state of worker or task $i$ \\
		\hline
		$\bm{h}^{t}(\bm{u}_i)$        &  hidden state of worker $\bm{u_i}$\\
		\hline
		$\bm{h}^{t}(\bm{v}_j)$        &  hidden state of task $\bm{v_j}$\\
		\hline
		$c_i$ & a constant coefficient \\\hline
		$\bm{W_r}$  & weight parameter used in MP1\\ \hline
		$\bm{W^u}, \bm{W^v}$  & weight parameters used in MP2       \\ \hline
		$\bm{W^u_e}, \bm{W^v_e} $  & weight parameters used in MP2          \\ \hline
		$\bm{W_1}, \bm{W_2} $  & weight parameters used in MP2         \\ \hline
		$b_1, b_2 $  & biases used in MP2          \\ \hline
		$\alpha_{ij}, \beta_{ij}$ & attention weights in MP2 \\ \hline
		$\bm{W^u_c}, \bm{W^v_c} $  &  weight parameters used in COR         \\ \hline
		$\bm{W_3}, \bm{W_4} $  & weight parameters used in COR \\ \hline
		$\gamma_{ij}, \delta_{ij} $  & attention weights in COR \\ \hline

	\end{tabular}
\end{table}

\section{Problem Statement and Notations}
In this paper, we study the crowdsourcing label aggregation problem. To formulate it, assume we have $n$ workers and $m$ tasks. The tasks can be classified into $K$ categories. For each task, a worker needs to select a single label out of $K$ candidate labels (we only consider the scenario of single-choice tasks, while a multi-choice task can be transformed into a set of single-choice tasks \cite{zheng2017truth,10.1145/2723372.2749430}). We denote the label that worker $i$ assigns to task $j$ as $l_{ij} \in \{1,...,K\}$. The goal of label aggregation in crowdsourcing is to infer the ground-truth label $y_j$ of each task $j$. In this work, we assume that we already have ground-truth labels for some tasks, and the task is to predict the remaining unknown labels for other tasks. Note that our method is applicable to both the case that each worker only assigns labels to part of the tasks and the case that each worker assigns labels to all of the tasks.

\section{Method}

In this section, we describe how our method is designed in detail.

\begin{figure}[h]
	\centering
	\includegraphics[width=0.5\linewidth]{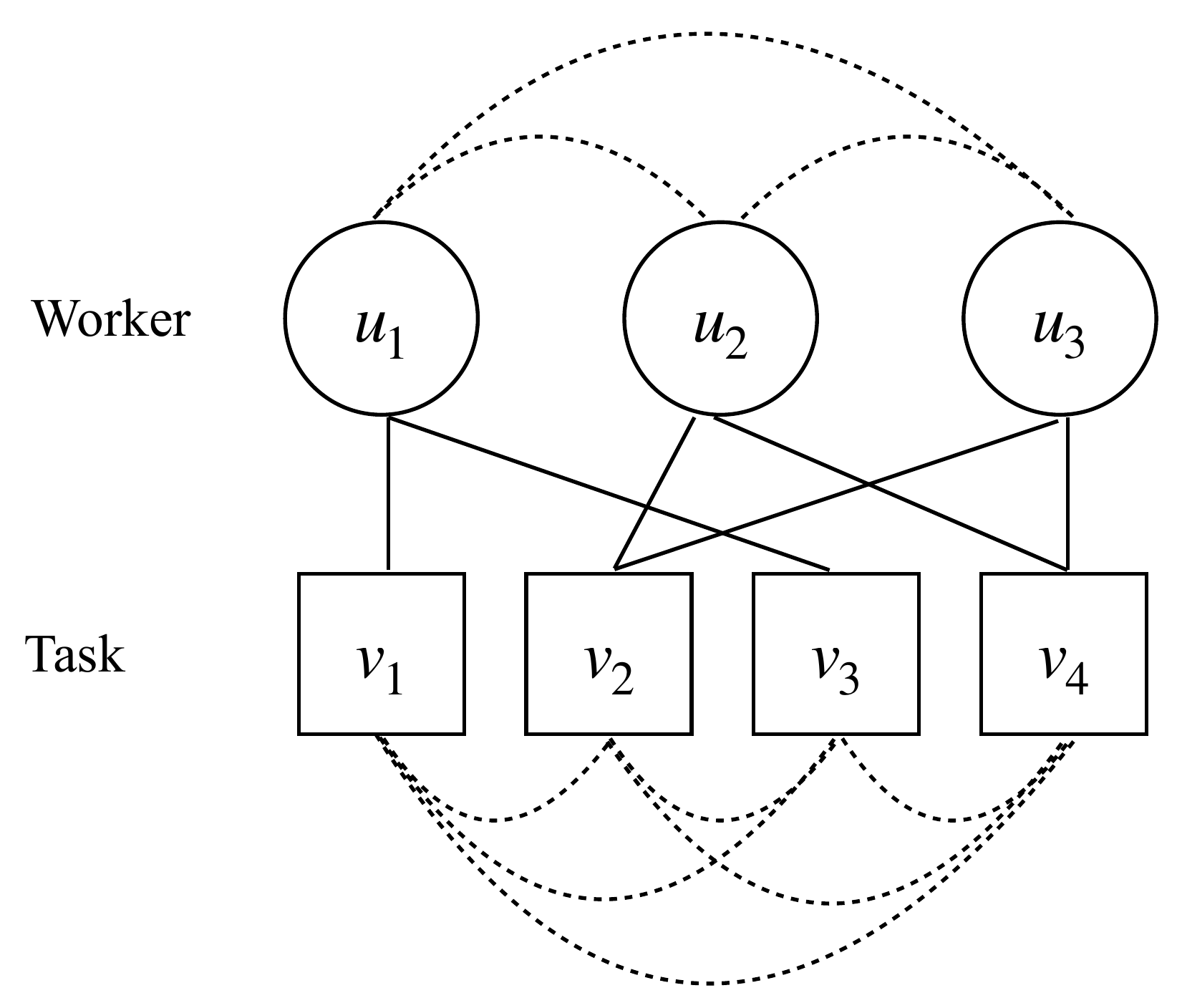}
	\caption{A worker-task assignment graph and the latent interaction between workers/tasks. $\bm{u}_i$ indicates the feature of the $i^{th}$ worker, and $\bm{v}_j$ indicates the feature of the $j^{th}$ task. Solid lines represent that a worker assigns a label to a task, while dashed lines represent the latent correlation between workers or between tasks. For simplicity, on the solid-line edges we omit the crowdsourced labels $l_{ij}$ that workers assign to tasks. }
	\label{fig:model}
\end{figure}

We first construct a graph to connect all the workers and tasks as shown in Fig.~\ref{fig:model}. Then we develop a new heterogeneous graph neural network to encode the worker nodes and task nodes into vector representations. Our new heterogeneous graph neural network contains two types of message passing layers \cite{gilmer2017neural}: the layer passing messages between workers and tasks, which captures the worker-task interactions; and the layer passing message among the same types of nodes, which captures the worker-worker correlation and task-task correlation.
After we get the node embeddings from the heterogeneous graph neural network, we add a prediction layer to predict the true label of each task.

\subsection{Motivation and Graph Construction}
Most previous methods for crowdsourcing formulate the label aggregation process as a complex generative process that is dependent on either worker ability or task difficulty. For example, in~\cite{whitehill2009whose},
\[
p(l_{ij} = z_j|\alpha_i,\beta_j) = \frac{1}{1+\exp{(-\alpha_i \beta_j})}
\]
where $l_{ij}$ is the label that worker $i$ assigned to task $j$, $z_j$ is the ground truth label of task $j$, $\alpha_i$ denotes the ability variable of worker $i$, and $\beta_j$ denotes the difficulty variable of task $j$. However, in these models we need to make delicate assumptions for the priors of these variables (e.g. Dirichlet priors) and carefully design a generative process, in order to make the inference tractable. In addition, the latent variables are generally scalars. This largely limits the modeling capacity because the worker's ability and task's difficulty may contain different aspects.

Inspired by the recent success of deep learning, we aim at using a deep neural network to explicitly learn the embeddings of worker features and task features which can determine the true labels. Considering that the labeling process can be represented as a graph, a graph neural network is a natural solution to the embedding problem. 

We show an example of the worker-task assignment graph in Fig.~\ref{fig:model}. In the graph, the nodes are either workers or tasks. If a worker $\bm{u}_i$ assigns a label to $\bm{v}_j$, there will be an edge connecting $\bm{u}_i$ and $\bm{v}_j$, and the edge feature is the one-hot crowdsourced label vector $\bm{e}_{ij} \in \{0,1\}^K$ which is derived from the label $l_{ij}$.

To initialize the features of nodes, we followed the feature representation method in~\cite{gaunt2016training}. We denote $g_j$ as the label of the task $u_i$ inferred by majority voting. For a worker node $u_i$, we calculate its features as below:
\begin{equation}
f(u_i) = \frac{|\left\{j\in \mathcal{N}(u_i) | l_{ij} = g_j\right\}|}{|\mathcal{N}(u_i)|}
\label{eq:feature_initial_worker}
\end{equation}
For a task node $v_j$,
\begin{equation}
f(v_j) = \frac{|\left\{i\in \mathcal{C}(v_j) | l_{ij} \not= g_j\right\}|}{|\mathcal{C}(v_j)|}
\label{eq:feature_initial_task}
\end{equation}
where $\mathcal{N}(u_i)$ denotes the set of tasks labeled by worker $u_i$ and $\mathcal{C}(v_j)$ is the set of workers that assigned labels to task $v_j$. $|*|$ denotes the cardinality of a set. This is based on an assumption that if the labels given by a worker is the same as the majority of people most of the time, he/she should have good labeling ability; for a task, the more worker who assigned different labels from the majority voting label to it, the more difficult the task can be. We fill the $d$-dimentional feature vector with the same value of $f(u_i)$ for worker $u_i$ and the same way for tasks. We also tried random initialization, the results can be found in Table~\ref{tab:feat-init}.

\begin{table}[h]
\caption{Comparison of Feature Initialization Methods.}
\label{tab:feat-init}
\begin{tabular}{|l|c|c|c|}
	\hline
	Datasets & Our Initialization Method       & Random Initialization      \\
	\hline
	bird      & \textbf{0.8610$\pm$0.0508} & 0.8517$\pm$0.0376 \\
	flowers   &\textbf{ 0.8638$\pm$0.0133} &0.8600$\pm$0.0169 \\
	web       & \textbf{0.9734$\pm$0.0215} &0.9284$\pm$0.0124\\
	dog       &\textbf{ 0.8243$\pm$0.0088} &0.8175$\pm$0.0098  \\
	rte       & \textbf{0.9269$\pm$0.0104} &0.9259$\pm$0.0103 \\
	SP        & \textbf{0.9149$\pm$0.0091} &0.9044$\pm$0.0045\\
	SP*       &\textbf{ 0.9445$\pm$0.0025} &0.9425$\pm$0.0040\\
	$\rm ZC_{all}$   & \textbf{0.9076$\pm$0.0162} &0.9012$\pm$0.0184 \\
	$\rm ZC_{in}$    & \textbf{0.7942$\pm$0.0071} &0.7828$\pm$0.0071\\
	$\rm ZC_{us}$    & \textbf{0.9130$\pm$0.0069} & 0.9034$\pm$0.0078\\
	face      & 0.6635$\pm$0.0118 & \textbf{0.6682$\pm$0.0126}\\
	product   & 0.9363$\pm$0.0019 &\textbf{0.9365$\pm$0.0023}\\
	sentiment & \textbf{0.9608$\pm$0.0076} & 0.9560$\pm$0.0060\\ \hline
\end{tabular}
\end{table}

\subsection{Message Passing Between Workers and Tasks}
Given the worker-task assignment graph, we cast the label aggregation problem as a node prediction problem in a heterogeneous graph neural network. To this aim, we develop a non-linear multi-layer message passing scheme for the graph node embedding. Message passing has been a key operation for many graph neural networks~\cite{zitnik2018modeling,gilmer2017neural}. The key idea is to propagate the information across all the edges of the graph in each layer. To illustrate, in the case of the worker-task graph, a worker's embedding is obviously impacted by its assigning labels and the corresponding tasks; and a task's embedding can also be inferred by the interaction with the workers who assign labels to it. In this paper, we implement two versions of message passing schemes between workers and tasks, denoted as MP1 and MP2 separately.

\subsubsection{MP1}

Following RGCN~\cite{schlichtkrull2018modeling}, one intuitive idea of message passing to update the hidden states of worker nodes and task nodes is the following formula, which we call MP1:
\begin{equation}
\bm{h}_i^{t+1} =\bm{h}_i^t + \frac{1}{|\mathcal{N}(i)|} \sum_{j \in \mathcal{N}(i)} \bm{W}_r \bm{h}_j^t
\label{eq:mp1_update}
\end{equation}
where $\mathcal{N}(i)$ represents the neighborhood of node $i$. When $i$ is a task node, $\mathcal{N}(i)$ denotes a set of workers that have assigned labels to it; When $i$ is a worker node, $\mathcal{N}(i)$ stands for a set of tasks that worker $i$ has assigned labels to. $\bm{W}_r$ is a matrix parameter for the edge label $l_{ij} = r$. In this way, we pass the message from workers to tasks and from tasks to workers.

\subsubsection{MP2}
The above message passing scheme assumes the neighbors have the same weight in the update function. This may lose importance information of different nodes. We can also employ the attention mechanism to re-weight the messages and derive another message passing scheme, MP2.

For a worker $\bm{u}_i$ with a hidden state $\bm{h}^t(\bm{u}_i)$, we update $\bm{h}^t(\bm{u}_i)$ by the following formula:
\begin{eqnarray}
\bm{h}^{t+1}(\bm{u}_i) &=& \phi\Big(c_i \bm{h}^t(\bm{u}_i) + (1-c_i) \\\nonumber
&& \sum_{\bm{v}_j \in \mathcal{N}(\bm{u}_i)} \alpha_{ij} M^u_t(\bm{h}^t(\bm{u}_i), \bm{h}^t(\bm{v}_j), \bm{e}_{ij})\Big)
\label{eq:mp2_worker_update}
\end{eqnarray}
where $M^u_t$ is the message function, $\phi$ is a nonlinear activation function (in this work we use ReLU), and $c_i \in [0,1]$ is a weight. 
In our case, the interaction between a worker and a task not only contains the worker/task node features, but also include the information of the crowdsourced labels. So our message function is calculated by taking into account both the node and the edge features. We first use a learnable matrix $\bm{W}^u_e$ to embed the edge vector $e_{ij}$ into an embedding vector and then concatenate it with the node features. Then we use an attention mechanism to re-weight the messages from different edges. 
\begin{equation}
M^u_t(\bm{h}^t(\bm{u}_i), \bm{h}^t(\bm{v}_j), \bm{e}_{ij}) =  \phi\Big( \bm{W}^u \left(\bm{h}^t(\bm{v}_j) \oplus (\bm{W}^u_e \bm{e}_{ij})\Big)\right)
\label{eq:message_u}
\end{equation}
where $\bm{W}^u$ is a parameter matrix, $\alpha_{ij}$ is the attention weight calculated by
\begin{equation}
\alpha_{ij} = \frac{\exp{(\bm{W_1} (M_{ij}\oplus \bm{h}^t(\bm{u}_i) + b_1) )}}{\sum_k \exp{(\bm{W_1} (M_{ik}\oplus \bm{h}^t(\bm{u}_i) + b_1))}}
\label{eq:alpha_ij}
\end{equation}
here we simplify $M^u_t(\bm{h}^t(\bm{u}_i), \bm{h}^t(\bm{v}_j), \bm{e}_{ij})$ as $M_{ij}$. 

Then we pass the messages from the workers to tasks. Similar to the above message passing phase, for each task $\bm{v}_j$, we also receive the messages from its connected edges and workers: 

\begin{equation}
M^v_t(\bm{h}^t(\bm{v}_j), \bm{h}^t(\bm{u}_i), \bm{e}_{ij}) =  \phi\Big( \bm{W}^v \left(\bm{h}^t(\bm{u}_i) \oplus (\bm{W}^v_e \bm{e}_{ij})\Big)\right)
\label{eq:message_v}
\end{equation}

We use a different matrix $\bm{W}^v_e$ for edge embedding, and a different parameter matrix $\bm{W}^v$. The attention weights are derived similarly by 
\begin{equation}
\beta_{ji} = \frac{\exp{(\bm{W_2} (M_{ji}\oplus \bm{h}^t(\bm{v}_j) + b_2) )}}{\sum_k \exp{(\bm{W_2} (M_{jk}\oplus \bm{h}^t(\bm{v}_j) + b_2))}}
\label{eq:beta_ij}
\end{equation}

Here we simplify $M^v_t(\bm{h}^t(\bm{v}_j), \bm{h}^t(\bm{u}_i), \bm{e}_{ij})$ as $M_{ji}$. Aggregating the messages from all the edges, we obtain the updated task embedding

\begin{eqnarray}
\bm{h}^{t+1}(\bm{v}_j) &=& \phi\Big(c_i \bm{h}^t(\bm{v}_j) + (1-c_i) \\\nonumber
&& \sum_{\bm{v}_j \in \mathcal{N}(\bm{u}_i)} \beta_{ji} M^v_t(\bm{h}^t(\bm{v}_j), \bm{h}^t(\bm{u}_i), \bm{e}_{ij})\Big)
\label{eq:mp2_task_update}
\end{eqnarray}

\subsection{COR: Latent Correlation Between Workers/Tasks}
The above message passing layer (either MP1 or MP2) explores the interaction between workers and tasks along the explicit edges which represent the assignment relationship. 
In practice, there may be also latent interaction/correlation among the same type of nodes (i.e. workers or tasks). For example, if two workers belong to the same community~\cite{venanzi2014community}, or they are close friends in a social network, they may have highly correlated preference or make similar mistakes in the labeling process. As to tasks, if their content is similar or they belong to the same category, it is highly possible that their labels have correlations. However, in a crowdsourcing platform, the explicit relationship among the workers or the tasks is often unknown. In this work, we develop a new layer to model the implicit inner-worker correlation and inner-task correlation and integrate the information into our new heterogeneous graph neural network. We denote this layer as COR.

Implicit worker correlation has been exploited in some Bayesian models before and demonstrated useful~\cite{venanzi2014community,cao2018max,li2019exploiting}. However, it is never explored in previous heterogeneous graph neural networks. Our model is also related to non-local neural networks~\cite{wang2018non} and self-attention models~\cite{vaswani2017attention}, which utilize long-range dependency of the inputs and improves the performance a lot.

Generally, a (heterogeneous) graph neural network requires to know the complete graph structures, i.e. all the edges. To utilize the correlation between the same type of nodes, we essentially add implicit edges among workers/tasks (based on some correlation function), as shown in Fig.~\ref{fig:model} (dashed lines).

Specifically in our model, for worker nodes, we assume that each node can be implicitly correlated to each of the other worker nodes. This is based on the assumption that even though two workers are not connected in the worker-task assignment graph (i.e. the two workers do not assign labels to the same task), they can still have some kind of implicit correlation between them.
But when we are faced with a quite large dataset, we can approximately reduce the number of neighbor nodes in the correlation layer to accelerate the message passing process. Two simple strategies are suggested, one is uniform sampling, the other is to select the 2-hop neighborhood in the worker-task assignment graph, i.e. only to capture the relations between two workers who share at least one task and between two tasks that are assigned to at least one same worker. Table~\ref{tab:node-sampling-comparison} shows the performance of our final model that using different neighborhood sampling strategies in the COR layer, both of the strategies have quite close performance to the original fully connected network. Inspired by \cite{velivckovic2017graph}, we update the worker embeddings as follows:

\begin{equation}
\bm{h}^{t+1}(\bm{u}_i) = \sigma \left(\sum_{\bm{u}_j \in \mathcal{N}}\gamma_{ij} \bm{W}_c^u \bm{h}^t(\bm{u}_j) \right)
\label{eq:cor_worker}
\end{equation}

where $\sigma$ is a non-linear activation function which is ReLU in our experiment. $\mathcal{N}$ denotes the set of all worker nodes including $\bm{u}_i$. $\bm{W}_c^u$ represents a parameter matrix. $\gamma_{ij}$ is the attention weight calculated by

\begin{equation}
\gamma_{i j}=\frac{\exp \left(\sigma\left(\bm{a}^{T}\left(\bm{W_3} \bm{h}^{t}\left(\bm{u}_{i}\right) \oplus \bm{W_3} \bm{h}^{t}\left(\bm{u}_{j}\right)\right)\right)\right)}{\sum_{\bm{u}_{k} \in \mathcal{N}} \exp \left(\sigma\left(\bm{a}^{T}\left(\bm{W_3} \bm{h}^{t}\left(\bm{u}_{i}\right) \oplus \bm{W_3} \bm{h}^{t}\left(\bm{u}_{k}\right)\right)\right)\right)}
\end{equation}
where $\bm{a}$ is a weight vector. We update the embeddings of task nodes in the same way as worker nodes, see the following equations. In our experiment, we found that only one head attention is enough for our task.

\begin{equation}
\bm{h}^{t+1}(\bm{v}_j) = \sigma \left(\sum_{\bm{v}_i \in \mathcal{C}}\delta_{ij} \bm{W}_c^v \bm{h}^t(\bm{v}_i) \right)
\label{eq:cor_task}
\end{equation}

\begin{equation}
\delta_{i j}=\frac{\exp \left(\sigma\left(\bm{b}^{T}\left(\bm{W_4} \bm{h}^{t}\left(\bm{v}_{j}\right) \oplus \bm{W_4} \bm{h}^{t}\left(\bm{v}_{i}\right)\right)\right)\right)}{\sum_{\bm{v}_{k} \in \mathcal{C}} \exp \left(\sigma\left(\bm{b}^{T}\left(\bm{W_4} \bm{h}^{t}\left(\bm{v}_{j}\right) \oplus \bm{W_4} \bm{h}^{t}\left(\bm{v}_{k}\right)\right)\right)\right)}
\end{equation}

\begin{table}[]
	\caption{A comparison between different neighbourhood sampling strategy.}
	\label{tab:node-sampling-comparison}
	\begin{tabular}{|l|c|c|c|}
		\hline
		Datasets & Fully Connected          & Uniform Sampling           & 2-Hop Neighbourhood         \\
		\hline
		bird      & 0.8610$\pm$0.0508 & 0.8402$\pm$0.0306 & 0.8449$\pm$0.0337 \\
		flowers   & 0.8638$\pm$0.0133 & 0.8688$\pm$0.0153 & 0.8638$\pm$0.0143 \\
		web       & 0.9734$\pm$0.0215 & 0.9703$\pm$0.0272 & 0.9852$\pm$0.0069 \\
		dog       & 0.8243$\pm$0.0088 & 0.8299$\pm$0.0101 & 0.8169$\pm$0.0138 \\
		rte       & 0.9269$\pm$0.0104 & 0.9263$\pm$0.0068 & 0.9284$\pm$0.0074 \\
		SP        & 0.9149$\pm$0.0091 & 0.9073$\pm$0.0081 & 0.9116$\pm$0.0080 \\
		SP*       & 0.9445$\pm$0.0025 & 0.9420$\pm$0.0033 & 0.9425$\pm$0.0025 \\
		$\rm ZC_{all}$   & 0.9076$\pm$0.0162 & 0.9006$\pm$0.0088 & 0.9050$\pm$0.0081 \\
		$\rm ZC_{in}$    & 0.7942$\pm$0.0071 & 0.7852$\pm$0.0031 & 0.7832$\pm$0.0096 \\
		$\rm ZC_{us}$    & 0.9130$\pm$0.0069 & 0.9062$\pm$0.0090 & 0.9022$\pm$0.0150 \\
		face      & 0.6635$\pm$0.0118 & 0.6665$\pm$0.0226 & 0.6670$\pm$0.0136 \\
		product   & 0.9363$\pm$0.0019 & 0.9354$\pm$0.0023 & 0.9351$\pm$0.0014 \\
		sentiment & 0.9608$\pm$0.0076 & 0.9588$\pm$0.0099 & 0.9583$\pm$0.0082 \\ \hline
	\end{tabular}
\end{table}

We analyze the complexity of our model in terms of each layer. We can split the edges into three categories: worker-worker, worker-task, task-task. Assume the worker-task edge set is $\mathcal{E}$, since we pass the messages from all these edges in MP1 layer, the complexity of MP1 layer is $O(|\mathcal{E}| d_t d_{t+1})$ where $d_t$ is the dimension of node embeddings at the $t^{th}$-layer.  The complexity of MP2 layer is $O(|\mathcal{E}| (d_t+d_e) d_{t+1})$ where $d_e$ is the dimension of the edge vector. The complexity of the correlation layer will be  $O((n^2+m^2)d_t d_{t+1})$. To reduce the complexity, we can use random sampling to sample only a subset of nodes as neighborhoods, or we can only use 2-hop neighborhoods in the correlation layer. As shown in table \ref{tab:node-sampling-comparison}, these approximations do not comprise much performance.

\subsection{Prediction and Training}
In previous sections, we introduced the message passing layer between workers and tasks, and the message passing layer between the same type of nodes. These layers can be stacked multiple times to get the final embeddings of workers and tasks. Then we can use the final task embeddings to predict their true labels. For a task $\bm{v}_j$ with the final embedding $\bm{h}(\bm{v}_j)$, we predict its label by:
\begin{equation}
\hat{\bm{y}}_j = \textrm{softmax}\big( \bm{W}_3  \bm{h}(\bm{v}_j) + b_3 \big)
\label{eq:predict_labels}
\end{equation}
We use the cross-entropy loss between the prediction $\hat{\bm{y}}_j (1\leq j \leq m)$ and the true labels $\bm{y}_j ( 1\leq j \leq m)$ as the loss function, 
\begin{equation}
L = \sum_{\bm{v_j}\in V_{train}, 1\leq k \leq K} y_{jk} \log \hat{y}_{jk} + (1-y_{jk}) \log (1-\hat{y}_{jk})
\label{eq:loss_function}
\end{equation}
where $y_{jk}$ and $\hat{y}_{jk}$ are the $k^{th}$ elements of $\bm{y}_j$ and $\hat{\bm{y}}_j$ separately. The model is then trained on the training tasks $V_{train}$ with known true labels with Adam and early stopping. The whole algorithm of our model MP2+COR+MP2 (i.e. stacked by an MP2 layer, a COR layer and another MP2 layer) can be expressed as below:

\begin{algorithm}
	\caption{MP2+COR+MP2}
	\begin{algorithmic}[1]
		\Require the worker-task assignment graph $G = (\mathcal{V}, \mathcal{E})$, where $\mathcal{V}$ consists of worker nodes $\mathcal{N}(u_i)$ and task nodes $\mathcal{C}(v_j)$. $\mathcal{E}$ is the set of edges between worker nodes and task nodes.  
		\Ensure the predicted true labels $y_j$ of each task nodes $v_j$. 
		\State Initialize the features of worker nodes $h_i^0$ by Equation~(\ref{eq:feature_initial_worker})  and the features of task nodes $h_j^0$ by Equation (\ref{eq:feature_initial_task}); Initialize the edge features as the one-hot label vector; t=1;
		\While {not converge and $t<t_{max}$ }
		\State Update  $h_i^t$ by Equation~(4) and $h_j^t$ by Equation~(9);
		\State Update worker features $h_i^t$ by Equation~(\ref{eq:cor_worker}), $h_j^t$ by Equation~(\ref{eq:cor_task});
		\State Update  $h_i^t$ by Equation~(4) and $h_j^t$ by Equation~(9);
		\State Predict the label $\hat{y}$ for tasks by Equation~(\ref{eq:predict_labels}).
		\State Obtain the loss by Equation~(\ref{eq:loss_function}) and update model parameters.
		\State t = t+1;
		\EndWhile
	\end{algorithmic}
\end{algorithm}

\section{Experiment}

\subsection{Datasets}
We ran our experiment on 13 widely-used real-world datasets.  These datasets are from four crowdsourcing dataset collections. Among them, \texttt{bird}, \texttt{dog}, \texttt{rte} and \texttt{web} are from 
\cite{zhang2014spectral} \footnote{https://github.com/zhangyuc/SpectralMethodsMeetEM},
\texttt{flowers} is obtained from \cite{tian2015uncovering}
\footnote{https://github.com/coverdark/deep\_laa}, \texttt{SP}, \texttt{SP*}, $\texttt{ZC}_{\texttt{all}}$, $\texttt{ZC}_{\texttt{in}}$ and $\texttt{ZC}_{\texttt{us}}$ are from \cite{venanzi2015activecrowdtoolkit} \footnote{https://github.com/orchidproject/active-crowd-toolkit}, \texttt{face}, \texttt{product} and \texttt{sentiment} are from \cite{zheng2017truth}
\footnote{https://zhydhkcws.github.io/crowd\_truth\_inference/index.html}.
Among them, ten datasets are binary tasks including \texttt{bird} to determine whether an image contains any bird \cite{welinder2010multidimensional}, \texttt{flowers} to distinguish whether the flower in an image is peach flower \cite{tian2015uncovering}, \texttt{rte} to recognize textual entailment \cite{snow2008cheap}, \texttt{SP} and \texttt{SP*} to perform sentiment analysis for movie reviews \cite{venanzi2015activecrowdtoolkit}, $\texttt{ZC}_{\texttt{all}}$, $\texttt{ZC}_{\texttt{in}}$, and $\texttt{ZC}_{\texttt{us}}$  to judge whether a URI is relevant to a named entity extracted from news \cite{venanzi2015activecrowdtoolkit}, \texttt{product} to tell whether two products are the same given their descriptions \cite{wang2012crowder}, \texttt{sentiment} to perform sentiment analysis for companies mentioned in tweets \cite{zheng2017truth}. There are also three multi-class tasks include \texttt{web} judging the relevance of web search results \cite{zhou2012learning}, \texttt{dog} determining the breed of a dog from ImageNet \cite{deng2009imagenet}, and \texttt{face} distinguishing the facial expressions \cite{mozafari2014scaling}.

	\begin{table}[h]
	\centering
	\caption{Datasets statistics}
	\label{dataset statistics}
	\scalebox{1}{
		\begin{tabular}{|l|r|r|c|r|} 
			\hline
			Dataset & \#Tasks & \#Workers &\#Categories & \#Labels\\
			\cline{1-5}
			bird & 108   & 39    & 2     & 4,212  \\  \cline{1-5} 
			flowers & 200   & 36    & 2     & 2,366  \\ \cline{1-5} 
			web & 2,653  & 177   & 5     & 15,539  \\ \cline{1-5} 
			dog & 807   & 109   & 4     & 8,070  \\ \cline{1-5} 
			rte & 800   & 164   & 2     & 8,000  \\ \cline{1-5} 
			SP & 4,999  & 203   & 2     & 27,746  \\ \cline{1-5} 
			SP* & 500   & 143   & 2     & 10,000  \\ \cline{1-5} 
			$\rm ZC_{all}$ & 2,040  & 78    & 2     & 20,125  \\ \cline{1-5} 
			$\rm ZC_{in}$ & 2,040  & 25    & 2     & 10,495  \\ \cline{1-5} 
			$\rm ZC_{us}$ & 2,040  & 74    & 2     & 11,155  \\ \cline{1-5} 
			face & 584   & 27    & 4     & 5,242  \\ \cline{1-5} 
			product & 8,315  & 176   & 2     & 24,945  \\ \cline{1-5} 
			sentiment & 1,000  & 85    & 2     & 20,000  \\
			\cline{1-5}
			\hline
		\end{tabular}
	}
\end{table}

The statistics of datasets are shown in Table~\ref{dataset statistics}, these datasets vary considerably in the number of tasks (from 108 to 8,315) and labels (from 2,366 to 27,746). The results of our experiments suggest that our method is adaptive to different scales of datasets. 
Our model has proven its capability of handling multi-label crowdsourcing problem by superior performance on these datasets. See Table~\ref{tab:Comparison with baselines}.

\subsection{Baselines}
We use these following methods for comparison:

\begin{itemize}
	\item \textbf{MV}: the MV is an abbreviation of majority voting, it is a basic model, which considers workers equally and selects the label that received most votes from workers as the true label. 
	
	\item \textbf{GLAD}:  the GLAD is an abbreviation of Generative model of Labels, Abilities, and Difficulties. This a probabilistic model that jointly infers the true label of each task, the expertise of workers, and the difficulty of tasks~\cite{whitehill2009whose}.                                                                                                                                                                                                 
	\item \textbf{MLP}: a three-layer MLP (Multi-Layer Perception) trained in a similar way as our method. 
	
	\item \textbf{iBCC-MF}: Bayesian Classifier Combination (BCC) was proposed by \cite{kim2012bayesian} for ensemble learning purpose. BCC has several variants,  iBCC-MF is a mean-field variational inference implementation of independent BCC (iBCC)~\cite{simpson2013dynamic, felt2015early, li2019exploiting} and performs slightly better than iBCC ~\cite{li2019exploiting}. Hence we include iBCC-MF as a baseline. 
	
	\item \textbf{EBCC}: an enhanced Bayesian classifier combination model proposed by~\citeauthor{li2019exploiting}~\cite{li2019exploiting}. This method models worker reliability at a subtype level, where each class is considered as a mixture of subtypes and worker performance at per subtype induces inter-worker correlations.
	
\end{itemize}

\subsection{Implementation Details}

Our model\footnote{https://github.com/whl97/Crowdsourcing\_Label\_Inference} is implemented based on Pytorch\footnote{https://pytorch.org} and Deep Graph Library (DGL)\footnote{http://dgl.ai}. We perform cross-validation to evaluate the performance of each model. Each dataset is separated into $n$ splits. We use one split for training and the rest for testing, and obtain the mean accuracy as the evaluation result. $n$ is set to 5, 10, and 20. Note that we randomly split the datasets and fix the splits afterward when evaluating all methods for a fair comparison.

\subsection{Results}
	\begin{table}
	\centering
	\caption{Accuracy comparison on 5-fold cross validation.}
	\scalebox{0.97}{
		\begin{tabular}{|l|cccccc|}
			\hline
			Dataset & MV & GLAD &MLP & iBCC-MF &EBCC & MP2+COR+MP2 \\
			\hline

			bird      & 0.7592 $\pm$0.0235 & 0.7593 $\pm$0.0149 & \textbf{0.9074$\pm$0.0218} & 0.8889 $\pm$0.0177 & 0.8610$\pm$0.0225  & 0.8610$\pm$0.0508 \\
			flowers   & 0.7600$\pm$0.0114 & 0.7950$\pm$0.0120 & 0.8213$\pm$0.0264 &\textbf{0.8700$\pm$0.0149} & 0.7200$\pm$0.0093 & 0.8638$\pm$0.0133 \\
			web       & 0.7765$\pm$0.0030 & 0.7252$\pm$0.0025 & 0.7982$\pm$0.0088 & 0.7508$\pm$0.0033 & 0.7437$\pm$ 0.0045 & \textbf{0.9734$\pm$0.0215} \\
			dog       & 0.8178$\pm$ 0.0052 & 0.8092$\pm$0.0054 & 0.6366$\pm$0.0117 & 0.8389$\pm$0.0050 & \textbf{0.8401$\pm$0.0057} & 0.8243$\pm$0.0088 \\
			rte       & 0.9188$\pm$0.0053 & 0.9050$\pm$0.0060 & 0.8463$\pm$0.0248 & 0.9275$\pm$0.0053 & \textbf{0.9313$\pm$0.0048} & 0.9269$\pm$0.0104 \\
			SP        & 0.8896$\pm$0.0018 & 0.8872$\pm$0.0013 & 0.8833$\pm$0.0114 & 0.9150$\pm$0.0019 & \textbf{0.9152$\pm$0.0017} & 0.9149$\pm$0.0091 \\
			SP*       & 0.9440$\pm$0.0034 & 0.9360$\pm$0.0034 & 0.9300$\pm$0.0132 & 0.9440$\pm$0.0034 &\textbf{0.9460$\pm$0.0022} &0.9445$\pm$0.0025\\
			$\rm ZC_{all}$  & 0.8348$\pm$0.0069 & 0.8294$\pm$0.0042 & 0.7936$\pm$0.0610 & 0.7951$\pm$0.0032 & 0.8632$\pm$0.0039 &\textbf{0.9076$\pm$0.0162} \\
			$\rm ZC_{in}$    & 0.7441$\pm$0.0013 & 0.7304$\pm$0.0020 & 0.7933$\pm$0.0154 & 0.7696$\pm$0.0034 & 0.7784$\pm$0.0039 & \textbf{0.7942$\pm$0.0071} \\
			$\rm ZC_{us}$    & 0.8696$\pm$0.0038 & 0.8221$\pm$0.0019 & 0.7830$\pm$0.0596 & 0.8270$\pm$0.0005 &0.9123$\pm$0.0023 & \textbf{0.9130$\pm$0.0069} \\
			face      & 0.6301$\pm$0.0102 & 0.6336$\pm$0.0086 & 0.6015$\pm$0.0156 & 0.6404$\pm$0.0082 & 0.6336$\pm$0.0062 & \textbf{0.6635$\pm$0.0118} \\
			product   & 0.8966$\pm$0.0020 & 0.9040$\pm$0.0016 & 0.8784$\pm$0.0017 & \textbf{0.9383$\pm$0.0012} & 0.9349$\pm$0.0016 & 0.9363$\pm$0.0019 \\
			sentiment & 0.9320$\pm$0.0038 & 0.9510$\pm$0.0046 & 0.9517$\pm$0.0048 & 0.9600$\pm$0.0055 &\textbf{0.9610$\pm$0.0045} &0.9608$\pm$0.0076 \\
			\hline

		\end{tabular}%
	}
	\label{tab:Comparison with baselines}%
\end{table}%

We compare our method with the aforementioned baselines on different real-world datasets. Table~\ref{tab:Comparison with baselines} compares the accuracy on different datasets under the 5-fold cross validation settings. The results demonstrate that our method outperforms others in most of the datasets. Due to the various natures of different datasets, it is hard for one crowdsourcing model to beat all others on all datasets (as shown in previous papers~\cite{li2019exploiting}). Among all 13 datasets, our method achieves the best accuracy on 5 datasets and is also comparable to the best performance on the other 8 datasets. The result on the dataset \texttt{web} is extremely remarkable, probably due to its good graph structure. When looking into detailed statistics of datasets, we notice that there are 7 datasets that have no less than 1000 tasks while other datasets are relatively small. Among the 7 larger datasets, our method achieves the highest accuracy on 4 of them and is less than 0.2\% worse than the best on the other 3 datasets. From another perspective, among 5 datasets on which we obtained the best results, 4 datasets are relatively larger. This suggests that our method is more superior on large datasets.

EBCC, another model with worker correlation in consideration, achieves the best results on 5 datasets (\texttt{dog}, \texttt{rte}, \texttt{SP}, \texttt{SP*}, and \texttt{sentiment}). Compared to EBCC, our method uses a different methodology from deep learning and graph neural networks, and achieves much more stable results across all datasets. Specifically, our model obtains the same accuracy on \texttt{bird}, and is better on 7 datasets (\texttt{flowers}, \texttt{web}, $\texttt{ZC}_{\texttt{all}}$ , $\texttt{ZC}_{\texttt{in}} $ ,  $\texttt{ZC}_{\texttt{us}}$, \texttt{face}, \texttt{product}, and only slightly inferior on 5 datasets (\texttt{dog}, \texttt{rte}, \texttt{SP}, \texttt{SP*} and \texttt{sentiment}).

It is worth noting that the MLP method has the same setting as our method, but the results are much worse than ours. That may be explained by the advantage of iterative message passing between workers and tasks in graph neural networks. Another reason may be that MLP can only utilize information from those tasks with ground truth during the training phase. Other tasks without ground-truth labels, however, have a lot of hidden information as well. Our method, as a semi-supervised graph neural network, is trained on the whole worker-task assignment graph, thus we can fully capture the hidden states of all tasks and workers and the structural information among them.

\subsection{Ablation Studies}

We study the effect of model components by comparing the prediction accuracy of different ablation models. Comparison of MP1, MP2 and their variants are shown in Table~\ref{prove_COR_layer}. MP$n$ $(n= 1, 2)$ denotes a single message passing layer, MP$n$+MP$n$ indicates that we stack two message passing layers, MP$n$+COR+MP$n$ means that we put a latent correlation layer between two message passing layers. The results show that on most of the datasets MP$n$+COR+MP$n$ almost constantly outperforms MP$n$+MP$n$ as well as the single layer MP$n$, regardless the selection of message passing method MP$n$. This demonstrates the effectiveness of capturing inter-worker and inter-task latent correlations. The COR layer brings in possible dependency between distant nodes, which the 2-hop model (MP$n$+MP$n$) cannot provide.

\begin{table}[h]
	\centering
	\caption{Prediction accuracy of different ablation model on 5-fold cross validation.}
	\label{prove_COR_layer}
	\scalebox{0.97}{
		
		\begin{tabular}{|l|ccc|ccc|}
			\hline
			Dataset &MP1 & MP1+MP1 &MP1+COR+MP1 &MP2 &MP2+MP2  & MP2+COR+MP2\\
			\hline
			bird           & 0.8472$\pm$0.0312 & 0.8219$\pm$0.0563 & 0.8658$\pm$0.0355 & 0.8841$\pm$0.0218 & 0.8609$\pm$0.0174 & 0.8610$\pm$0.0508 \\
			flowers        & 0.8163$\pm$0.0140 & 0.8287$\pm$0.0191 & 0.8438$\pm$0.0288 & 0.8475$\pm$0.0230 & 0.8575$\pm$0.0163 & 0.8638$\pm$0.0133 \\
			web            & 0.8585$\pm$0.0076 & 0.8606$\pm$0.0062 & 0.9428$\pm$0.0232 & 0.9509$\pm$0.0065 & 0.9710$\pm$0.0056 & 0.9734$\pm$0.0215 \\
			dog            & 0.8324$\pm$0.0077 & 0.8271$\pm$0.0069 & 0.8278$\pm$0.0074 & 0.8206$\pm$0.0161 & 0.8042$\pm$0.0146 & 0.8243$\pm$0.0088 \\
			rte            & 0.9256$\pm$0.0062 & 0.9256$\pm$0.0085 & 0.9272$\pm$0.0099 & 0.9284$\pm$0.0050 & 0.9269$\pm$0.0072 & 0.9269$\pm$0.0104 \\
			SP             & 0.8971$\pm$0.0037 & 0.9019$\pm$0.0059 & 0.9113$\pm$0.0032 & 0.9130$\pm$0.0026 & 0.9138$\pm$0.0032 & 0.9149$\pm$0.0091 \\
			SP*            & 0.9455$\pm$0.0052 & 0.9440$\pm$0.0057 & 0.9435$\pm$0.0021 & 0.9425$\pm$0.0029 & 0.9420$\pm$0.0021 & 0.9445$\pm$0.0025 \\
			$\rm ZC_{all}$ & 0.8456$\pm$0.0093 & 0.8513$\pm$0.0078 & 0.8739$\pm$0.0062 & 0.8989$\pm$0.0042 & 0.9083$\pm$0.0034 & 0.9076$\pm$0.0162 \\
			$\rm ZC_{in}$  & 0.7828$\pm$0.0071 & 0.7828$\pm$0.0071 & 0.7828$\pm$0.0071 & 0.7875$\pm$0.0023 & 0.7904$\pm$0.0059 & 0.7942$\pm$0.0071 \\
			$\rm ZC_{us}$  & 0.8757$\pm$0.0085 & 0.8795$\pm$0.0042 & 0.8819$\pm$0.0076 & 0.8968$\pm$0.0047 & 0.9062$\pm$0.0057 & 0.9130$\pm$0.0069 \\
			face           & 0.6678$\pm$0.0127 & 0.6712$\pm$0.0151 & 0.6742$\pm$0.0182 & 0.6675$\pm$0.0102 & 0.6618$\pm$0.0205 & 0.6635$\pm$0.0118 \\
			product        & 0.9232$\pm$0.0009 & 0.9295$\pm$0.0020 & 0.9314$\pm$0.0015 & 0.9336$\pm$0.0026 & 0.9338$\pm$0.0024 & 0.9363$\pm$0.0019 \\
			sentiment      & 0.9535$\pm$0.0067 & 0.9500$\pm$0.0091 & 0.9530$\pm$0.0095 & 0.9563$\pm$0.0059 & 0.9545$\pm$0.0084 & 0.9608$\pm$0.0076 \\
			\hline
		\end{tabular}%
	}
\end{table}%

\begin{figure}[h]
	\centering
	\includegraphics[width=0.5\linewidth]{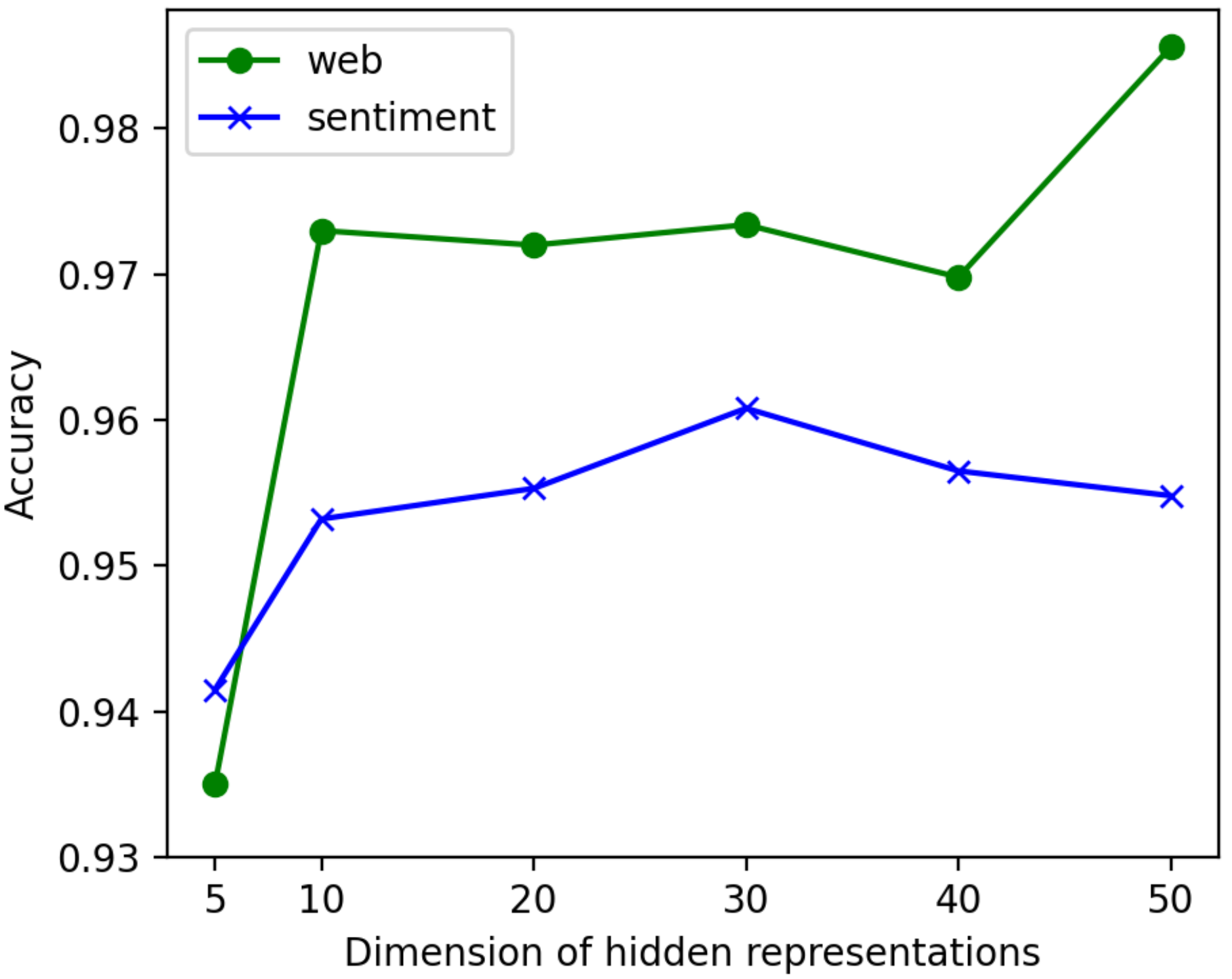}
	
	\caption{Effect of different dimensions of hidden representations. We display the accuracies of our final model (MP2-COR-MP2) on two datasets, \texttt{web} and \texttt{sentiment}, along the change of dimensionality.}
	\label{fig:model_sensitivity}
\end{figure}

\subsection{Effect of Dimensionality}

We also study the impact of the dimensions of hidden representations. We experiment on the proposed MP2+COR+MP2 model. As shown in Fig.~\ref{fig:model_sensitivity},  the best dimension for each dataset to obtain the highest accuracy are not always the same. When faced with a new dataset, it is difficult for us to know the best dimension. Thus we fix this hyperparameter to 30 for all datasets to present the final results.

\subsection{Effect of Training Proportion}

\begin{figure*}
	\centering
	\includegraphics[width=\linewidth]{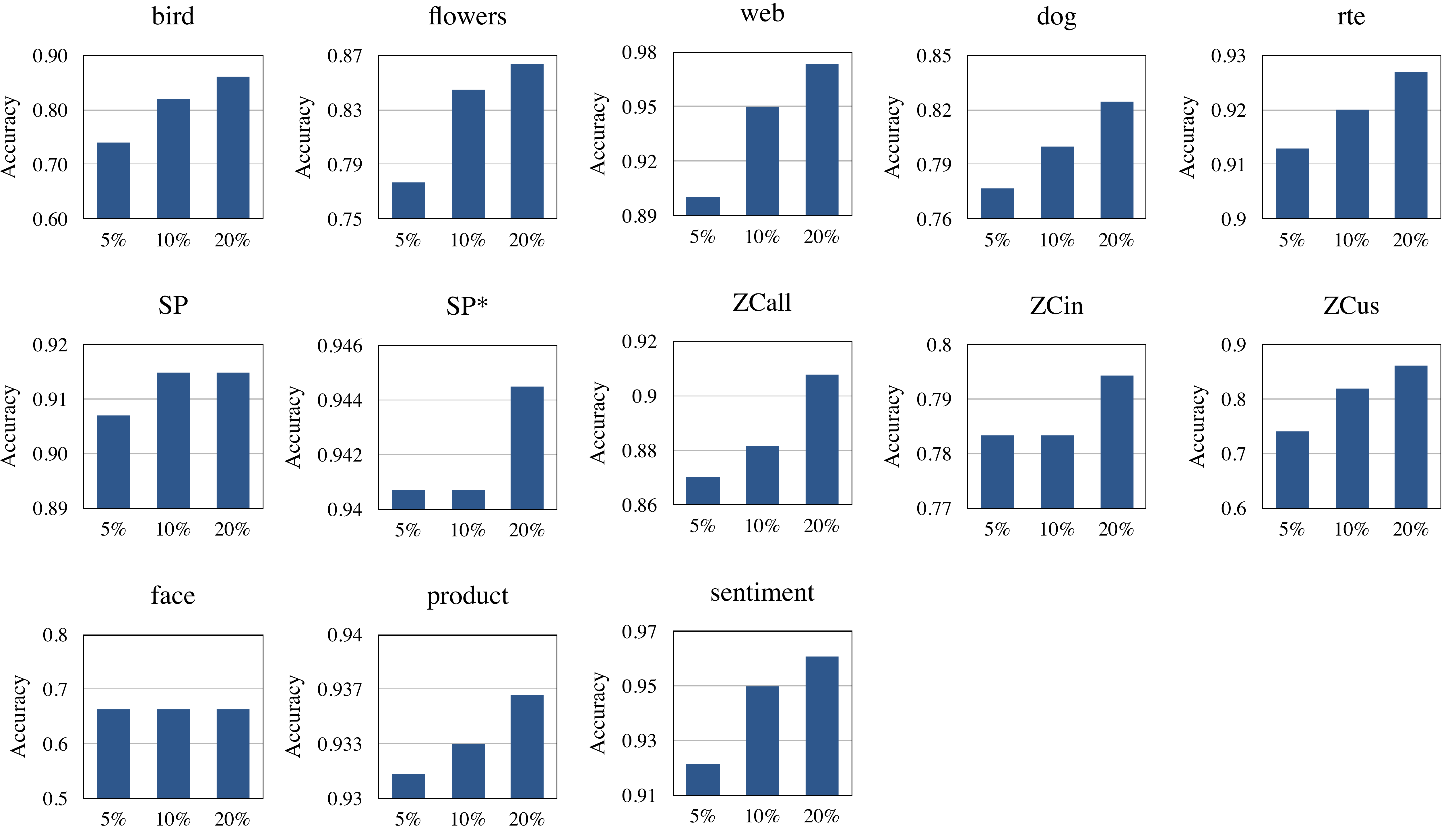}
	
	\caption{Effect of training proportion on different datasets. Each subgraph shows the accuracies of our final model (MP2-COR-MP2) on different datasets along the change of training proportion (5\%, 10\%, and 20\%). }
	\label{fig:train_proportion_comparisom}
\end{figure*}

To study the effect of different training proportions, Fig.~\ref{fig:train_proportion_comparisom} demonstrates how the performance of our model varies with the training proportion on each dataset. On all datasets, the accuracy increases as the training proportion becomes larger. But the trends of some datasets are barely noticeable, which indicates that on these datasets our method can achieve quite good performance with very little training data (e.g. 5\%). Some other datasets increase obviously with the proportion of training data, we find that our model can fully utilize the training data and achieve quite remarkable performance compared to other methods (e.g. on \texttt{web} and $\texttt{ZC}_{\texttt{all}}$).

\section{Conclusion and Future work}

We present a novel Heterogeneous Graph Neural Network for label aggregation in crowdsourcing. Constructing a graph to represent the worker-task interactions, we utilize the power of graph neural networks to learn a better representation for workers and tasks. Moreover, our heterogeneous graph neural network differs from previous works by adding new latent correlations among the same type of nodes (i.e. worker nodes and task nodes), which captures the worker-worker and task-task correlation in the crowdsourcing problem. Comparing with state-of-the-art label aggregation models and our own ablation models, we demonstrated the effectiveness of heterogeneous graph neural networks on real-world crowdsourcing datasets, as well as the usefulness of modeling the latent correlation of workers/tasks. Future work includes exploring the generative models for crowdsourcing graphs and extends our model to the unsupervised setting (without the requirement of ground-truth labels).

\bibliographystyle{ACM-Reference-Format}
\bibliography{acmart.bib}

\end{document}